\documentclass[runningheads]{llncs}
\usepackage{graphicx}
\usepackage{comment}
\usepackage{amsmath,amssymb} 
\usepackage{color}

\usepackage{gensymb} 
\usepackage{threeparttable}
\usepackage{framed}
\usepackage{booktabs}
\usepackage{multirow}
\usepackage{extarrows}

\begin{document}

\pagestyle{headings}
\mainmatter
\def\ECCVSubNumber{3539}  

\title{3D-Rotation-Equivariant Quaternion Neural Networks} 

\titlerunning{3D-Rotation-Equivariant Quaternion Neural Networks}
\authorrunning{W. Shen et al.}

{
	\renewcommand{\thefootnote}{\fnsymbol{footnote}}
	
	\author{Wen Shen$^{2,*}$ \and
		Binbin Zhang$^{2,*}$ \and
		Shikun Huang$^{2,*}$ \and
		Zhihua Wei$^{2}$ \and\\
		Quanshi Zhang$^{1,\dag}$}
	
	\institute{$^1$Shanghai Jiao Tong University,\qquad $^2$Tongji University, Shanghai
		\email{\{wen\_shen,0206zbb,hsk,zhihua\_wei\}@tongji.edu.cn,zqs1022@sjtu.edu.cn}\\
	}
	
	\footnotetext[1]{Wen Shen, Binbin Zhang and Shikun Huang have equal contributions. This study was done when they worked as interns at SJTU.}
	\footnotetext[4]{Quanshi Zhang is the corresponding author and is with the John Hopcroft Center and MoE Key Lab of Artificial Intelligence AI Institute, Shanghai Jiao Tong University, Shanghai, China.}
}
\maketitle

\begin{abstract}
This paper proposes a set of rules to revise various neural networks for 3D point cloud processing to rotation-equivariant quaternion neural networks (REQNNs). We find that when a neural network uses quaternion features, the network feature naturally has the rotation-equivariance property. Rotation equivariance means that applying a specific rotation transformation to the input point cloud is equivalent to applying the same rotation transformation to all intermediate-layer quaternion features. Besides, the REQNN also ensures that the intermediate-layer features are invariant to the permutation of input points. Compared with the original neural network, the REQNN exhibits higher rotation robustness.
\end{abstract}

\section{Introduction}
3D point cloud processing has attracted increasing research attention in recent years. Unlike images with rich color information, 3D point clouds mainly use spatial contexts for feature extraction. Therefore, the rotation is not supposed to have essential impacts on 3D tasks, such as 3D shape classification and reconstruction. Besides, reordering input points should not have crucial effects on these tasks as well, which is termed the permutation-invariance property.

In this study, we focus on the problem of learning neural networks for 3D point cloud processing with rotation equivariance and permutation invariance.

\begin{figure}[tbp]
	\begin{center}
		\resizebox{\linewidth}{!}{
		\includegraphics[width=\linewidth]{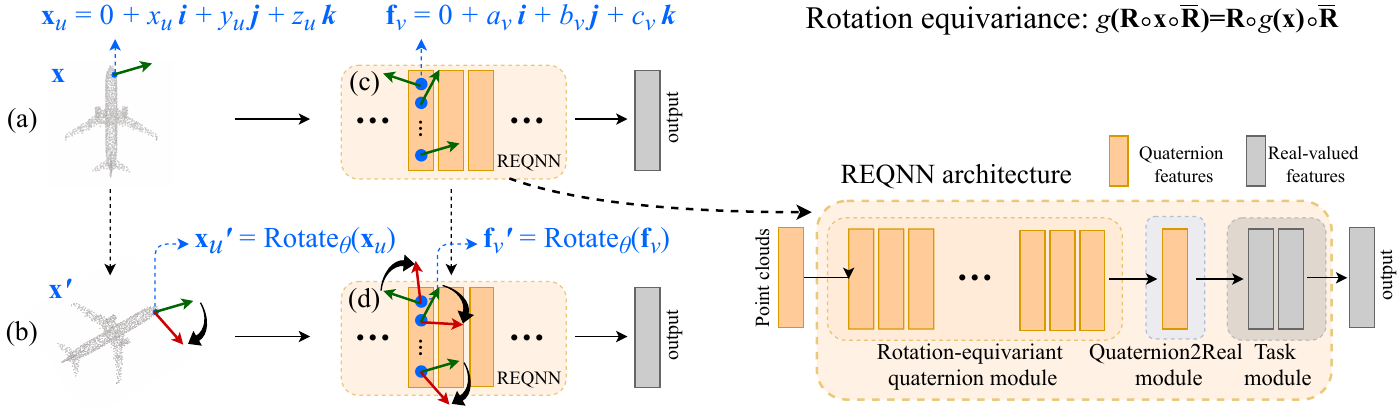}
	}
	\end{center}
	\caption{\textbf{Overview.} In the REQNN, both input point clouds and intermediate-layer features are represented by quaternion features (blue). In this paper, the rotation equivariance is defined as follows. When we rotate the input point cloud $\mathbf{x}$ (a) with a specific angle (\emph{e.g.} 60\degree) to obtain the same point cloud $\mathbf{x}'$ (b) with a different orientation, then the intermediate-layer feature (d) generated by the REQNN is equivalent to applying the transformation \emph{w.r.t.} the same rotation to the feature (c) of the original point cloud. \emph{I.e.} the rotation equivariance is defined as $g({\rm Rotate}_\theta(\mathbf{x}))={\rm Rotate}_\theta(g(\mathbf{x}))$. REQNNs exhibit significantly higher rotation robustness than traditional neural networks.}
	\label{fig:figure1}
\end{figure}

\noindent
$\bullet$ \textbf{Rotation equivariance:} Rotation equivariance has been discussed in recent research \cite{cohen2016steerable}. In this study, we define rotation equivariance for neural networks as follows. If an input point cloud is rotated by a specific angle, then the feature generated by the network is equivalent to applying the transformation \emph{w.r.t.} the same rotation to the feature of the original point cloud (see Fig.~\ref{fig:figure1} (left)). In this way, we can use the feature of a specific point cloud to synthesize features of the same point cloud with different orientations. Specifically, we can apply the transformation of a specific rotation to the current feature to synthesize the target feature.

\begin{sloppypar}
\noindent
$\bullet$ \textbf{Permutation invariance:} Permutation invariance measures whether intermediate-layer features essentially keep unchanged when we reorder input 3D points. Fortunately, we find that quaternion features in a neural network naturally satisfy the rotation-equivariance property under certain conditions (details will be introduced later). Therefore, we propose a set of rules to revise most existing neural networks to rotation-equivariant quaternion neural networks (REQNNs). Given a specific neural network for 3D point cloud processing (\emph{e.g.} PointNet \cite{qi2017pointnet}, PointNet++ \cite{qi2017pointnet++}, DGCNN \cite{wang2018dynamic}, PointConv \cite{wu2019pointconv}, etc., which are learned for various tasks, such as 3D shape classification and reconstruction), our rules can help revise the network to a REQNN with both properties of rotation equivariance and permutation invariance.
\end{sloppypar}

To revise a neural network to a REQNN with rotation equivariance, we transform both the input and intermediate-layer features of the original neural network into quaternion features (\emph{i.e.} vectors/matrices/tensors, in which each element is a quaternion). A quaternion is a hyper-complex number with three imaginary parts ($\boldsymbol{i},\boldsymbol{j}$, and $\boldsymbol{k}$) \cite{hamilton1848xi}. 3D rotations can be represented using quaternions. \emph{I.e.} rotating a quaternion $\mathbf{q}\in\mathbb{H}$ with an angle $\theta\in[0,2\pi)$ around an axis $\mathbf{o}=0+o_1\boldsymbol{i}+o_2\boldsymbol{j}+o_3\boldsymbol{k}\in\mathbb{H}$ ($o_1$, $o_2$, $o_3\in\mathbb{R}$) can be represented as $\mathbf{R}\mathbf{q}\overline{\mathbf{R}}$, where $\mathbf{R}=e^{\mathbf{o}\frac{\theta}{2}}\in\mathbb{H}$; $\overline{\mathbf{R}}=e^{-\mathbf{o}\frac{\theta}{2}}\in\mathbb{H}$ is the conjugation of $\mathbf{R}$.

In this way, the rotation equivariance of a REQNN is defined as follows. When we apply a specific rotation to the input {$\mathbf{x}\in\mathbb{H}^n$}, \emph{i.e.} {$\mathbf{x}'\!\!=\!\!\mathbf{R}\circ\mathbf{x}\circ\overline{\mathbf{R}}$}, the network will generate an intermediate-layer quaternion feature {$g(\mathbf{x}')\in\mathbb{H}^d$}, where $\circ$ denotes the element-wise multiplication. The rotation equivariance ensures that {$g(\mathbf{x}')=\mathbf{R}\circ g(\mathbf{x})\circ\overline{\mathbf{R}}$}. Note that the input and the feature here can be vectors/matrices/tensors, in which each element is a quaternion.

Therefore, we revise a number of layerwise operations in the original neural network to make them rotation-equivariant, such as operations of the convolution, ReLU, batch-normalization, max-pooling, 3D coordinates weighting \cite{wu2019pointconv}, etc., in order to ensure the rotation-equivariance property.

Note that most tasks, such as the shape classification, require outputs composed of real numbers. However, the REQNN's features consist of quaternions. Therefore, for real applications, we revise quaternion features of a specific high layer of the REQNN into ordinary features, in which each element is a real number. We transform quaternion features into real numbers by using the square of the norm of each quaternion element in the feature to replace the corresponding feature element. We put the revised real-valued features into the last few layers to generate real-valued outputs, as Fig.~\ref{fig:figure1} (right) shows. Such revision ensures that the last few layerwise operations are rotation invariant. We will introduce this revision in Section \ref{sec:quaternion_to_real}.

Besides the rotation-equivariance property, the REQNN is also supposed to have the permutation-invariance property as follows. When we reorder 3D points in the input point cloud {$\mathbf{x}$} to obtain the same point cloud $\mathbf{x}^{\rm reorder}$ with a different order, the network will generate the same feature, \emph{i.e.} {$g(\mathbf{x}^{\rm reorder})=g(\mathbf{x})$}. Therefore, we revise a few operations in the original neural network to be permutation invariant, \emph{e.g.} the farthest point sampling \cite{qi2017pointnet++} and the ball-query-search-based grouping \cite{qi2017pointnet++}, to ensure the permutation-invariance property of the REQNN.

In this study, we do not limit our attention to a specific architecture. Our method can be applied to various neural networks for different tasks. Experimental results proved that REQNNs exhibited superior performance than the original networks in terms of rotation robustness.

Contributions of our study are summarized as follows. We propose a set of generic rules to revise various neural networks to REQNNs with both properties of rotation equivariance and permutation invariance. The proposed rules can be broadly applied to different neural networks for different tasks, such as 3D shape classification and point cloud reconstruction. Experiments have demonstrated the effectiveness of our method that REQNNs exhibit higher rotation robustness than traditional neural networks.

\section{Related Work}

\textbf{Deep learning for 3D point cloud processing:} Recently, a series of studies have focused on deep neural networks (DNNs) for 3D point cloud processing and have achieved superior performance in various 3D tasks \cite{qi2017pointnet,su2018splatnet,simonovsky2017dynamic,yu2018pu,wang2018sgpn,shi2019pointrcnn}. As a pioneer of using DNNs for 3D point cloud processing, PointNet \cite{qi2017pointnet} aggregated all individual point features into a global feature using a max-pooling operation. In order to further extract contextual information of 3D point clouds, existing studies have made lots of efforts. PointNet++ \cite{qi2017pointnet++} hierarchically used PointNet as a local descriptor. KC-Net \cite{shen2018mining} proposed kernel correlation to measure the similarity between two point sets, so as to represent local geometric structures around each point. PointSIFT \cite{jiang2018pointsift} proposed a SIFT-like operation to encode contextual information of different orientations for each point. Point2Sequence \cite{liu2019point2sequence} employed an RNN-based encoder-decoder structure to capture correlations between different areas in a local region by aggregating multi-scale areas of each local region with attention.

Unlike images, 3D point clouds cannot be processed by traditional convolution operators. To address this problem, Kd-network \cite{klokov2017escape} built a kd-tree on subdivisions of the point cloud, and used such kd-tree structure to mimics the convolution operator to extract and aggregate features according to the subdivisions. PointCNN \cite{li2018pointcnn} proposed an $\mathcal{X}$-Conv operator to aggregate features from neighborhoods into fewer representative points. Pointwise CNN \cite{hua2018pointwise} binned nearest neighbors into kernel cells of each point and convolved them with kernel weights. PointConv \cite{wu2019pointconv} treated convolution kernels as nonlinear functions that were learned from local coordinates of 3D points and their densities, respectively. Besides, some studies introduced graph convolutional neural networks for the extraction of geodesic information \cite{simonovsky2017dynamic,wang2018dynamic}. Some studies focused on the use of spatial relations between neighboring points \cite{liu2019relation,zhao2019pointweb}. In this study, we aim to learn DNNs with properties of rotation equivariance and permutation invariance.

\textbf{3D rotation robustness:} The most widely used method to improve the rotation robustness was data augmentation \cite{van2001art}. However, data augmentation significantly boosted the computational cost. Spatial Transformer Networks (STNs) \cite{jaderberg2015spatial} allowed spatial manipulations of data and features within the network, which improved the rotation robustness.

Some studies went beyond rotation robustness and focused on rotation invariance. The rotation-invariance property means that the output always keeps unchanged when we rotate the input. One intuitive way to achieve rotation invariance was to project 3D points onto a sphere \cite{you2018prin,rao2019spherical,zhang2019new} and constructed spherical CNNs \cite{s.2018spherical} to extract rotation-invariant features. Other studies learned rotation-invariant representations that discarded orientation information of input point clouds \cite{chen2019clusternet,deng2018ppf,zhang2019rotation}.

However, such rotation-invariant methods directly discarded rotation information, so the rotation equivariance is proposed as a more promising property of feature representations. Rotation-equivariant methods both encode rotation information and disentangle rotation-independent information from the point cloud. To the best of our knowledge, there were very few studies in this direction. Previous studies developed specific network architectures \cite{zhao2019quaternion} or designed specific operations \cite{thomas2018tensor} to achieve rotation equivariance. In comparison, we aim to propose a set of generic rules to revise most existing neural networks to achieve the rotation-equivariance property. Unlike \cite{zhao2019quaternion,thomas2018tensor}, our method can be applied to various neural networks for different tasks.

\textbf{Complex and quaternion networks:} Recently, besides neural networks using real-valued features, neural networks using complex-valued features or quaternion-valued \cite{hamilton1848xi} features have been developed \cite{arjovsky2016unitary,wisdom2016full,danihelka2016associative,wolter2018complex,guberman2016complex,trabelsi2017deep,xiang2019complex,gaudet2018deep,kendall2015posenet,zhu2018quaternion,parcollet2018quaternion}. In this study, we use quaternions to represent intermediate-layer features and 3D rotations to achieve 3D rotation-equivariance property.

\section{Approach}

\subsection{Quaternion Features in Neural Networks and Rotations}\label{sec:quaternion_nn}

\begin{framed}
	\textbf{Quaternion:} A quaternion \cite{hamilton1848xi} {$\mathbf{q}=q_0+q_1\boldsymbol{i}+q_2\boldsymbol{j}+q_3\boldsymbol{k}\in\mathbb{H}$} is a hyper-complex number with a real part ({$q_0$}) and three imaginary parts ({$q_1\boldsymbol{i},q_2\boldsymbol{j},q_3\boldsymbol{k}$}), where {$q_0,q_1,q_2,q_3\in\mathbb{R}$}; $\mathbb{H}$ denotes the algebra of quaternions. If the real part of {$\mathbf{q}$} is 0, then {$\mathbf{q}$} is a \textbf{pure quaternion}. If the norm of a quaternion {$\lVert \mathbf{q} \rVert=\sqrt{q_0^2+q_1^2+q_2^2+q_3^2}=1$}, then {$\mathbf{q}$} is a \textbf{unit quaternion}. The \textbf{conjugation} of $\mathbf{q}$ is $\overline{\mathbf{q}}=q_0-q_1\boldsymbol{i}-q_2\boldsymbol{j}-q_3\boldsymbol{k}$.
	
	The products of basis elements $\boldsymbol{i}$, $\boldsymbol{j}$, and $\boldsymbol{k}$ are defined by $\boldsymbol{i}^2=\boldsymbol{j}^2=\boldsymbol{k}^2=\boldsymbol{i}\boldsymbol{j}\boldsymbol{k}=-1$ and $\boldsymbol{i}\boldsymbol{j}=\boldsymbol{k}, \boldsymbol{j}\boldsymbol{k}=\boldsymbol{i},\boldsymbol{k}\boldsymbol{i}=\boldsymbol{j},\boldsymbol{j}\boldsymbol{i}=-\boldsymbol{k},\boldsymbol{k}\boldsymbol{j}=-\boldsymbol{i}$, and $\boldsymbol{i}\boldsymbol{k}=-\boldsymbol{j}$. Note that the multiplication of two quaternions is non-commutative, \emph{i.e.} $\boldsymbol{i}\boldsymbol{j}\neq \boldsymbol{j}\boldsymbol{i},\boldsymbol{j}\boldsymbol{k}\neq \boldsymbol{k}\boldsymbol{j}$, and $\boldsymbol{k}\boldsymbol{i}\neq \boldsymbol{i}\boldsymbol{k}$.
	
	Each quaternion has a \textbf{polar decomposition}. In this study, we only focus on the polar decomposition of a unit quaternion in the form of $\mathbf{q}=\cos\frac{\theta}{2}+\sin\frac{\theta}{2}(o_1\boldsymbol{i}+o_2\boldsymbol{j}+o_3\boldsymbol{k})$, $\sqrt{o_1^2+o_2^2+o_3^2}=1$. The polar decomposition of such a unit quaternion is $\mathbf{q}=e^{\mathbf{o}\frac{\theta}{2}}$, where $\mathbf{o}=o_1\boldsymbol{i}+o_2\boldsymbol{j}+o_3\boldsymbol{k}$. As aforementioned, multiplication of two quaternions is non-commutative, therefore, $e^{\mathbf{o}\frac{\theta}{2}}\mathbf{p}e^{-\mathbf{o}\frac{\theta}{2}}\neq \mathbf{p}$.
\end{framed}

For a traditional neural network, inputs, features, and parameters are vectors/matrices/tensors, in which each element is a real number. However, in a REQNN, inputs and features are vectors/matrices/tensors composed of quaternions; parameters are still vectors/matrices/tensors composed of real numbers.

\begin{sloppypar}
\textbf{Quaternion inputs and features:} In a REQNN, each $u$-th point
($[x_u,y_u,z_u]^{\top}\in\mathbb{R}^3$) in a 3D point cloud is represented as a pure quaternion $\mathbf{x}_u=0+x_u\boldsymbol{i}+y_u\boldsymbol{j}+z_u\boldsymbol{k}$. Each $v$-th element of the intermediate-layer feature is also a pure quaternion $\mathbf{f}_v=0+a_v\boldsymbol{i}+b_v\boldsymbol{j}+c_v\boldsymbol{k}$, where {$a_v,b_v,c_v\in\mathbb{R}$}.
\end{sloppypar}

\textbf{Quaternion rotations:} Each element of a feature, $\mathbf{f}_v=0+a_v\boldsymbol{i}+b_v\boldsymbol{j}+c_v\boldsymbol{k}$, can be considered to have an orientation, \emph{i.e.} $[a_v,b_v,c_v]^{\top}$. In this way, 3D rotations can be represented using quaternions. Suppose we rotate $\mathbf{f}_v$ around an axis {$\mathbf{o}=0+o_1\boldsymbol{i}+o_2\boldsymbol{j}+o_3\boldsymbol{k}$} (where {$o_1,o_2,o_3\in\mathbb{R}$, $\lVert\mathbf{o}\rVert=1$}) with an angle {$\theta\in[0,2\pi)$} to get $\mathbf{f}'_v$. Such a rotation can be represented using a unit quaternion $\mathbf{R}=\cos\frac{\theta}{2}+\sin\frac{\theta}{2}(o_1\boldsymbol{i}+o_2\boldsymbol{j}+o_3\boldsymbol{k})=e^{\mathbf{o}\frac{\theta}{2}}$ and its conjugation $\overline{\mathbf{R}}$, as follows.
\begin{equation}
\mathbf{f}'_v=\mathbf{R}\mathbf{f}_v\overline{\mathbf{R}}.
\end{equation}
Note that {$\mathbf{f}'_v=\mathbf{R}\mathbf{f}_v\overline{\mathbf{R}}\neq\mathbf{f}_v$}. The advantage of using quaternions to represent rotations is that quaternions do not suffer from the singularity problem, but the Euler Angle \cite{weisstein2009euler} and the Rodrigues parameters \cite{shuster1993survey} do. Besides, although the redundancy ratio of quaternions is two, the redundancy does not affect the rotation equivariance property of the REQNN.

To ensure that all quaternion features are rotation equivariant, all imaginary parts (\emph{i.e.} $\boldsymbol{i}$, $\boldsymbol{j}$, and $\boldsymbol{k}$) of a quaternion element share the same \textbf{real-valued} parameter $w$. Take the convolution operation $\otimes$ as an example, $w\otimes \mathbf{f}=w\otimes (0+a\boldsymbol{i}+b\boldsymbol{j}+c\boldsymbol{k})= 0+(w\otimes a)\boldsymbol{i} + (w\otimes b)\boldsymbol{j}  + (w\otimes c)\boldsymbol{k} $, where $w$ is the \textbf{real-valued} parameter; $\mathbf{f}$ is the quaternion feature; $a$, $b$, and $c$ are \textbf{real-valued} tensors of the same size for the convolution operation in this example.

\subsection{Rotation Equivariance}
In order to recursively achieve the rotation-equivariance property for a REQNN, we should ensure that each layerwise operation of the REQNN has the rotation-equivariance property. In a REQNN, the rotation equivariance is defined as follows. Let {$\mathbf{x}\in\mathbb{H}^{n}$} and {$\mathbf{y}=\mathbf{\Phi}(\mathbf{x})\in\mathbb{H}^C$} denote the input and the output of the REQNN, respectively. Note that outputs for most tasks are traditional vectors/matrices/tensors, in which each element is a real number. In this way, we learn rotation-equivariant quaternion features in most layers, and then transform these features into ordinary real-valued rotation-invariant features in the last few layers, as shown in Fig.~\ref{fig:figure1} (right). We will introduce details for such revision in Section \ref{sec:quaternion_to_real}.

For each rotation $\mathbf{R}=e^{\mathbf{o}\frac{\theta}{2}}$ and its conjugation {$\overline{\mathbf{R}}$}, the rotation equivariance of a REQNN is defined as follows.
\begin{equation}\label{eq:reqnn}
\mathbf{\Phi}(\mathbf{x}^{(\theta)}) =  \mathbf{R}\circ \mathbf{\Phi}(\mathbf{x})\circ\overline{\mathbf{R}},\quad {\rm s.t.}\quad \mathbf{x}^{(\theta)}\triangleq \mathbf{R}\circ \mathbf{x}\circ\overline{\mathbf{R}},
\end{equation}
where $\circ$ denotes the element-wise multiplication (\emph{e.g.} $\mathbf{x}^{(\theta)}\triangleq \mathbf{R}\circ \mathbf{x}\circ\overline{\mathbf{R}}$ can also be formulated as $\mathbf{x}_u^{(\theta)}\triangleq \mathbf{R}\mathbf{x}_u\overline{\mathbf{R}}, u=1,2,...,n$). As discussed in the previous paragraph, outputs for most tasks are real-valued features. Therefore, Equation~(\ref{eq:reqnn}) does not hold for all layers in the neural network. Instead, we transform features in last few layers to be real-valued rotation-invariant features.

To achieve the above rotation equivariance, we must ensure the layerwise rotation equivariance. Let {$\mathbf{\Phi}(\mathbf{x})=\Phi_L(\Phi_{L-1}(\cdots\Phi_1(\mathbf{x}))) $} represent the cascaded functions of multiple layers of a neural network, where {$\Phi_l(\cdot)$} denotes the function of the {$l$}-th layer. Let {$\mathbf{f}_l=\Phi_l(\mathbf{f}_{l-1})\in\mathbb{H}^d$} denote the output of the {$l$}-th layer. The layerwise rotation equivariance is given as follows.
\begin{equation}\label{eq:layerwise_eq}
\Phi_l(\mathbf{f}_{l-1}^{(\theta)})=\mathbf{R}\circ \Phi_l(\mathbf{f}_{l-1})\circ\overline{\mathbf{R}},\ \  {\rm s.t.}\ \  \mathbf{f}_{l-1}^{(\theta)}\triangleq \mathbf{R}\circ \mathbf{f}_{l-1}\circ\overline{\mathbf{R}}.
\end{equation}
This equation recursively ensures the rotation-equivariance property of the REQNN. Let us take a neural network with three layers as a toy example. $\mathbf{\Phi}(\mathbf{x}^{\theta}) =\Phi_3(\Phi_2(\Phi_1(\mathbf{R}\circ\mathbf{x}\circ\overline{\mathbf{R}})))=\Phi_3(\Phi_2(\mathbf{R}\circ \Phi_1(\mathbf{x})\circ\overline{\mathbf{R}})) =\Phi_3(\mathbf{R}\circ\Phi_2( \Phi_1(\mathbf{x}))\circ\overline{\mathbf{R}})
	=\mathbf{R}\circ\Phi_3(\Phi_2(\Phi_1(\mathbf{x})))\circ\overline{\mathbf{R}}
	=\mathbf{R}\circ \mathbf{\Phi}(\mathbf{x}) \circ\overline{\mathbf{R}}$. Please see supplementary materials for more discussion.

\begin{table}[tbp]
	\begin{center}
		\resizebox{\linewidth}{!}{
			\begin{tabular}{l|c|c|l|c|c}
				\toprule[1pt]
				\multicolumn{1}{c|}{\multirow{2}{*}{Operation}}& Rotation & Permutation &\multicolumn{1}{c|}{\multirow{2}{*}{Operation}}& Rotation& Permutation\\
				& equivariance& invariance && equivariance& invariance \\
				\hline\hline
				Convolution &$\times$ &-- &Grouping ($k$-NN)\cite{wu2019pointconv} & \checkmark & \checkmark \\
				ReLU & $\times$&-- &Grouping (ball query) \cite{qi2017pointnet++} & \checkmark & $\times$ \\
				Batch-normalization &$\times$ & --&Density estimation \cite{wu2019pointconv} & \checkmark & \checkmark\\
				Max-pooling &$\times$ &-- &3D coordinates weighting \cite{wu2019pointconv} & $\times$ & \checkmark\\
				Dropout  & $\times$ & -- &Graph construction \cite{wang2018dynamic} & \checkmark &  \checkmark\\
				Farthest point sampling\cite{qi2017pointnet++} & \checkmark & $\times$ & & &\\
				\bottomrule[1pt]
			\end{tabular}	
		}
	\end{center}
	\caption{Rotation-equivariance and permutation-invariance properties of layerwise operations in the original neural network. ``$\times$'' denotes that the operation does not have the property, ``$\checkmark$'' denotes that the operation naturally has the property, and ``--'' denotes that the layerwise operation is naturally unrelated to the property (which will be discussed in the last paragraph of Section \ref{sec:rules_of_permutation}). Please see Section~\ref{sec:rules_of_rotation} and Section~\ref{sec:rules_of_permutation} for rules of revising layerwise operations to be rotation-equivariant and permutation-invariant, respectively.}
	\label{tab:rules}
\end{table}

\subsection{Rules for Rotation Equivariance}\label{sec:rules_of_rotation}
We propose a set of rules to revise layerwise operations in the original neural network to make them rotation-equivariant, \emph{i.e.} satisfying Equation~(\ref{eq:layerwise_eq}). Table~\ref{tab:rules} shows the list of layerwise operations in the original neural network with the rotation-equivariance property, and those without the rotation-equivariance property. \emph{The rotation-equivariance property of the revised layerwise operations has been proved in supplementary materials.}

\textbf{Convolution:} We revise the operation of the convolution layer, {$Conv(\mathbf{f})=w\otimes \mathbf{f} +b$}, to be rotation-equivariant by removing the bias term {$b$}, where $w$ is the real-valued parameter and $\mathbf{f}$ is the quaternion feature.

\begin{sloppypar}
\textbf{ReLU:} We revise the ReLU operation as follows to make it rotation-equivariant.
\end{sloppypar}
\begin{equation}\label{eq:relu}
ReLU(\mathbf{f}_v) = \frac{\lVert\mathbf{f}_v\rVert}{\max\{\lVert\mathbf{f}_v\rVert,c\}} \mathbf{f}_v,
\end{equation}
where $\mathbf{f}_v\in\mathbb{H}$ denotes the $v$-th element in the feature $\mathbf{f}\in\mathbb{H}^d$; $c$ is a positive constant, which can be implemented as $c=\frac{1}{d}\sum_{v=1}^{d} \lVert\mathbf{f}_v\rVert$.

\textbf{Batch-normalization:} We revise the batch-normalization operation to be rotation-equivariant, as follows.
\begin{equation}\label{eq:norm}
norm(\mathbf{f}_v^{(i)})=\frac{\mathbf{f}_v^{(i)}}{ \sqrt{\mathbb{E}_j[\lVert \mathbf{f}_v^{(j)}\rVert^2]+ \epsilon}},
\end{equation}
where $\mathbf{f}^{(i)}\in\mathbb{H}^d$ denotes the feature of the $i$-th sample in the batch; $\epsilon$ is a tiny positive constant to avoid dividing by 0.

\textbf{Max-pooling:} We revise the max-pooling operation, as follows.
\begin{equation}\label{eq:maxpooling}
\textit{maxPool}({\bf f})= {\bf f}_{\hat{v}}\quad{\rm s.t.}\quad \hat{v}=\mathop{\arg\max}_{v=1,\dots,d.}[\Vert{\bf f}_v\Vert].
\end{equation}
Note that for 3D point cloud processing, a special element-wise max-pooling operation designed in \cite{qi2017pointnet} is widely used. The revision for this special max-pooling operation can be decomposed to a group of operations as Equation~(\ref{eq:maxpooling}) defined. Please see our supplementary materials for revision details of this operation.

\textbf{Dropout:} For the dropout operation, we randomly drop out a number of quaternion elements from the feature. For each dropped element, both the real and imaginary parts are set to zero. Such revision naturally satisfies the rotation-equivariance property in Equation~(\ref{eq:layerwise_eq}).

\textbf{3D coordinates weighting:} The 3D coordinates weighting designed in \cite{wu2019pointconv} focuses on the use of 3D coordinates' information to reweight intermediate-layer features. This operation is not rotation-equivariant, because the rotation changes coordinates of points. To make this operation rotation-equivariant, we use the Principal Components Analysis (PCA) to transform 3D points to a new local reference frame (LRF). Specifically, we choose eigenvectors corresponding to the first three principal components as new axes ${\rm x}$, ${\rm y}$, and ${\rm z}$ of the new LRF. In this way, the coordinate system rotates together with input points, so the transformed new coordinates are not changed. Note that contrary to \cite{zhao2019quaternion} relying on the LRF, our research only uses LRF to revise the 3D coordinates weighting operation, so as to ensure the specific neural network designed in \cite{wu2019pointconv} to be rotation equivariant.

The following five layerwise operations in the original neural network, which are implemented based on distances between points, are naturally rotation-equivariant, including the farthest point sampling \cite{qi2017pointnet++}, the $k$-NN-search-based grouping \cite{wang2018dynamic,wu2019pointconv}, the ball-query-search-based grouping \cite{qi2017pointnet++}, the density estimation \cite{wu2019pointconv}, and the graph construction \cite{wang2018dynamic} operations.

\subsection{Rules for Permutation Invariance}\label{sec:rules_of_permutation}

As shown in Table~\ref{tab:rules}, the farthest point sampling \cite{qi2017pointnet++}, and the ball-query-search-based grouping \cite{qi2017pointnet++} are not permutation-invariant. Therefore, we revise these two operations to be permutation-invariant as follows.

\textbf{Farthest point sampling:} The farthest point sampling (FPS) is an operation for selecting a subset of points from the input point cloud, in order to extract local features \cite{qi2017pointnet++}. Suppose that we aim to select $n$ points from the input point cloud, if $i-1$ points have already been selected, \emph{i.e.} {$S_{i-1}=\{{x}_1,x_2,\dots,x_{i-1}\}$}, then the next selected point {$x_{i}$} is the farthest point from {$S_{i-1}$}. The FPS is not permutation-invariant, because the subset selected by this operation depends on which point is selected first. To revise the FPS to be permutation-invariant, we always use the centroid of a point cloud, which is a virtual point, as the first selected point. In this way, the FPS would be permutation-invariant.

\textbf{Grouping (ball query):} The ball-query-search-based grouping is used to find $K$ neighboring points within a radius for each given center point, in order to extract contextual information \cite{qi2017pointnet++}. This operation is not permutation-invariant, because when there are more than $K$ points within the radius, the top $K$ points will be selected according to the order of points. To revise this operation to be permutation-invariant, we replace the ball query search by $k$-NN search when the number of points within the radius exceeds the required number.

Other operations that implemented based on distances between points are permutation-invariant, because reordering input points has no effects on distances between points, including the $k$-NN-search-based grouping \cite{wang2018dynamic,wu2019pointconv}, the density estimation \cite{wu2019pointconv}, the 3D coordinates weighting \cite{wu2019pointconv}, and the graph construction \cite{wang2018dynamic} operations.

Note that there is no need to discuss the permutation invariance of the convolution, the ReLU, the batch-normalization, the max-pooling, and the dropout operations. It is because the permutation invariance of these operations depends on receptive fields. \emph{I.e.} if the receptive field of each neural unit keeps the same when we reorder input points, then the operation is permutation-invariant. Whereas receptive fields are determined by other operations (\emph{e.g.} the FPS and grouping).

\subsection{Overview of the REQNN}\label{sec:quaternion_to_real}
Although using quaternions to represent intermediate-layer features helps achieve the rotation-equivariance property, most existing tasks (\emph{e.g.} the shape classification) require outputs of real numbers. Thus, we need to transform quaternion features into ordinary real-valued features, in which each element is a real number. Note that for the point cloud reconstruction task, features of the entire neural network are quaternions. It is because outputs required by the point cloud reconstruction task are 3D coordinates, which can be represented by quaternions.

Therefore, as Fig.~\ref{fig:figure1} shows, the REQNN consists of (a) rotation-equivariant quaternion module, (b) Quaternion2Real module, and (c) task module.

\begin{sloppypar}
\textbf{Rotation-equivariant quaternion module:} Except for very few layers on the top of the REQNN, other layers in the REQNN comprise the rotation-equivariant quaternion module. This module is used to extract rotation-equivariant quaternion features. We use rules proposed in Section \ref{sec:rules_of_rotation} to revise layerwise operations in the original neural network to be rotation-equivariant, so as to obtain the rotation-equivariant quaternion module. We also use rules in Section \ref{sec:rules_of_permutation} to revise these layerwise operations to be permutation invariant.
\end{sloppypar}

\textbf{Quaternion2Real module:} The Quaternion2Real module is located after the rotation-equivariant quaternion module. The Quaternion2Real module is used to transform quaternion features into real-valued vectors/matrices/tensors as features. Specifically, we use an element-wise operation to compute the square of the norm of each quaternion element as the real-valued feature element. \emph{I.e.} for each $v$-th element of a quaternion feature, $\mathbf{f}_v=0+a_v\boldsymbol{i}+b_v\boldsymbol{j}+c_v\boldsymbol{k}$, we compute the square of the norm {$\lVert \mathbf{f}_v \rVert^2 ={a_v^2+b_v^2+c_v^2}$} as the corresponding element of the real-valued feature. Note that the transformed features are rotation-invariant.

\textbf{Task module:} The task module is composed of the last few layers of the REQNN. The task module is used to obtain ordinary real-valued outputs, which are required by the task of 3D shape classification. As aforementioned, the Quaternion2Real module transforms quaternion features into real-valued vectors/matrices/tensors as features. In this way, the task module (\emph{i.e.} the last few layers) in the REQNN implements various tasks just like traditional neural networks.

\begin{table*}[t]
	\begin{center}
		\resizebox{\linewidth}{!}{
			\begin{tabular}{l|c|c|c|c|c|c|c|c}
				\toprule[1pt]
				&\multicolumn{2}{c|}{PointNet++\footnotemark[1] \cite{qi2017pointnet++}}& \multicolumn{2}{c|}{DGCNN\footnotemark[2] \cite{wang2018dynamic}} & \multicolumn{2}{c|}{PointConv \cite{wu2019pointconv}} &\multicolumn{2}{c}{PointNet \cite{qi2017pointnet}}  \\
				\cline{2-3}\cline{4-5}\cline{6-7}\cline{8-9}
				& FLOPs(G)  & $\#$Params(M) & FLOPs(G)  & $\#$Params(M)   & FLOPs(G)  & Params(M) & FLOPs(G)  & $\#$Params(M)   \\
				\hline\hline
				Ori.& 0.87 &1.48& 3.53 & 2.86 & 1.44& 19.57&0.30&0.29\\
				\hline
				REQNN &2.51 &1.47 & 8.24 & 2.86 & 4.22 &20.61&0.88&0.28\\
				\bottomrule[1pt]
		\end{tabular}	}
	\end{center}
	\caption{Comparisons of the number of floating-point operations (FLOPs) and the number of parameters ($\#$Params) of original neural networks and REQNNs. All neural networks were tested on the ModelNet40 dataset.}
	\label{tab:complexity}
\end{table*}

\begin{table*}[tbp]
	\begin{center}
		\resizebox{\linewidth}{!}{
			\begin{tabular}{l|c|c|c|c}
				\toprule[1pt]
				\multicolumn{1}{c|}{Layerwise operation} & PointNet++ \cite{qi2017pointnet++}& DGCNN \cite{wang2018dynamic} & PointConv \cite{wu2019pointconv} &PointNet \cite{qi2017pointnet}\\
				\hline\hline
				Convolution &\checkmark&\checkmark  & \checkmark &  \checkmark \\
				ReLU & \checkmark&\checkmark  & \checkmark & \checkmark\\
				Batch-normalization &\checkmark & \checkmark & \checkmark &  \checkmark\\
				Max-pooling &\checkmark & \checkmark &  &  \checkmark\\
				Dropout  & \checkmark &\checkmark&\checkmark&\checkmark \\
				Farthest point sampling & \checkmark & & \checkmark  &\\
				Grouping ($k$-NN) &  & \checkmark & \checkmark & \\
				Grouping (ball query) \cite{qi2017pointnet++} & \checkmark & &  & \\
				Density estimation \cite{wu2019pointconv} &  & & \checkmark & \\
				3D coordinates weighting \cite{wu2019pointconv} & & & \checkmark & \\
				Graph construction \cite{wang2018dynamic} &  &  \checkmark & & \\
				\bottomrule[1pt]
			\end{tabular}	
		}
	\end{center}
	\caption{Layerwise operations of different neural networks. ``$\checkmark$'' denotes that the network contains the layerwise operation.}
	\label{tab:nn}
\end{table*}

\textbf{Complexity of the REQNN:} The REQNN’s parameter number is no more than that of the original neural network. The REQNN's operation number is theoretically less than three times of that of the original neural network. We have compared numbers of operations and numbers of parameters of original neural networks and REQNNs in Table~\ref{tab:complexity}.

\subsection{Revisions of Traditional Neural Networks into REQNNs}\label{sec:reqnn}

In this study, we revise the following four neural networks to REQNNs, including PointNet++ \cite{qi2017pointnet++}, DGCNN \cite{wang2018dynamic}, PointConv \cite{wang2018dynamic}, and PointNet \cite{qi2017pointnet}.

\textbf{Model 1, PointNet++:} As Table~\ref{tab:nn} shows, the PointNet++ \cite{qi2017pointnet++} for shape classification includes seven types of layerwise operations. To revise the PointNet++ for shape classification\footnote[1]{The PointNet++ for shape classification used in this paper is slightly revised by concatenating 3D coordinates to input features of the 1-st and 4-th convolution layers, in order to enrich the input information. For fair comparisons, both the REQNN and the original PointNet++ are revised in this way.} to a REQNN, we take the last three fully-connected (FC) layers as the task module and take other layers as the rotation-equivariant quaternion module. We add a Quaternion2Real module between these two modules. We use rules proposed in Section~\ref{sec:rules_of_rotation} to revise four types of layerwise operations to be rotation-equivariant, including the convolution, ReLU, batch-normalization, and max-pooling operations. We also use rules proposed in Section \ref{sec:rules_of_permutation} to revise farthest point sampling and ball-query-search-based grouping operations in the original PointNet++ to be permutation-invariant.

\textbf{Model 2, DGCNN:} As Table~\ref{tab:nn} shows, the DGCNN \cite{wang2018dynamic} for shape classification contains seven types of layerwise operations. To revise the DGCNN for shape classification to a REQNN, we take the last three FC layers as the task module and take other layers as the rotation-equivariant quaternion module. The Quaternion2Real module\footnote[2]{We add one more convolution layer in the Quaternion2Real module in the REQNN revised from DGCNN, in order to obtain reliable real-valued features considering that the DGCNN has no downsampling operations. For fair comparisons, we add the same convolution layer to the same location of the original DGCNN.} is added between these two modules. We revise four types of layerwise operations to be rotation-equivariant, including the convolution, ReLU, batch-normalization, and max-pooling operations. All layerwise operations in the original DGCNN are naturally permutation-invariant. Therefore, there is no revision for permutation invariance here.

\textbf{Model 3, PointConv:} As Table~\ref{tab:nn} shows, the PointConv \cite{wu2019pointconv} for shape classification includes eight types of layerwise operations. To revise the PointConv for shape classification to a REQNN, we take the last three FC layers as the task module and take other layers as the rotation-equivariant quaternion module. The Quaternion2Real module is added between these two modules. We revise the following four types of layerwise operations to be rotation-equivariant, \emph{i.e.} the convolution, ReLU, batch-normalization, and 3D coordinates weighting operations. We also revise all farthest point sampling operations in the original PointConv to be permutation-invariant.

\textbf{Model 4, PointNet:} In order to construct a REQNN for shape reconstruction, we slightly revise the architecture of the PointNet \cite{qi2017pointnet} for shape classification. As Table~\ref{tab:nn} shows, the PointNet for shape classification contains five types of layerwise operations. We take all remaing layers in the PointNet as the rotation-equivariant quaternion module except for the max-pooling and Spatial Transformer Network (STN) \cite{jaderberg2015spatial}. The STN discards all spatial information (including the rotation information) of the input point cloud. Therefore, in order to encode rotation information, we remove the STN from the original PointNet.

Note that there is no the Quaternion2Real module or the task module in this REQNN, so that all features in the REQNN for reconstruction are quaternion features. We revise the following four types of layerwise operations to be rotation-equivariant, \emph{i.e.} the convolution, the ReLU, the batch-normalization, and the dropout operations.

\begin{table*}[tbp]
	\begin{center}
		\resizebox{\linewidth}{!}{
			\begin{tabular}{l|c|c|c|c|c|c}
				\toprule[1pt]
				\multicolumn{1}{c|}{\multirow{3}{*}{Method}}&\multicolumn{3}{c|}{ModelNet40 dataset}& \multicolumn{3}{c}{3D MNIST dataset}\\
				\cline{2-4}\cline{5-7}
				& Baseline & Baseline & \multirow{2}{*}{REQNN} & Baseline  & Baseline & \multirow{2}{*}{REQNN}  \\
				& w/o rotations & w/ rotations&   & w/o rotations  & w/ rotations&   \\
				\hline\hline
				PointNet++\footnotemark[1] \cite{qi2017pointnet++} & 23.57\footnotemark[3] &26.43 & \textbf{63.95} & 44.15 & 51.16 & \textbf{68.99}\\
				\hline
				DGCNN\footnotemark[2] \cite{wang2018dynamic}  &30.05\footnotemark[3]  &31.34 & \textbf{83.03}& 45.37& 49.25 &\textbf{82.09}\\
				\hline
				PointConv \cite{wu2019pointconv} & 21.93 &23.72  & \textbf{78.14}  &44.63 &50.95&\textbf{78.59}\\
				\bottomrule[1pt]
		\end{tabular}	}
	\end{center}
	\caption{Accuracy of 3D shape classification on the ModelNet40 and the 3D MNIST datasets. ``Baseline w/o rotations'' indicates the original neural network learned without rotations. ``Baseline w/ rotations'' indicates the original neural network learned with the ${\rm z}$-axis rotations (data augmentation with the ${\rm z}$-axis rotations has been widely applied in \cite{qi2017pointnet++,wang2018dynamic,wu2019pointconv}). ``REQNN'' indicates the REQNN learned without rotations. Note that the accuracy of shape classification reported in \cite{qi2017pointnet++,wang2018dynamic,wu2019pointconv} was obtained under the test without rotations. The accuracy reported here was obtained under the test with rotations. Therefore, it is normal that the accuracy in this paper is lower than the accuracy in those papers.}
	\label{tab:results_class}
\end{table*}

\begin{table}[t]
	\newcommand{\tabincell}[2]{\begin{tabular}{@{}#1@{}}#2\end{tabular}}
	\begin{center}
		\resizebox{0.95\linewidth}{!}{
			\begin{tabular}{l|c|c}
				\toprule[1pt]
				\multicolumn{1}{c|}{\multirow{2}{*}{Method}}& \textbf{NR/NR} (do \textbf{not} & \textbf{NR/AR} \\
				&  consider rotation in testing)& (consider rotation in testing)\\
				\hline\hline
				PointNet \cite{qi2017pointnet} & 88.45 & 12.47\\
				PointNet++ \cite{qi2017pointnet++} & 89.82 &21.35\footnotemark[3]   \\
				Point2Sequence \cite{liu2019point2sequence} & 92.60 & 10.53\\
				KD-Network \cite{klokov2017escape}  &86.20 & 8.49   \\
				RS-CNN \cite{liu2019relation} &92.38 &22.49 \\
				DGCNN \cite{wang2018dynamic} & \textbf{92.90} & 29.74\footnotemark[3] \\
				PRIN \cite{you2018prin} & 80.13 &68.85 \\
				QE-Capsule network \cite{zhao2019quaternion}  & 74.73&74.07 \\
				\hline
				\tabincell{l}{REQNN \\ (revised from DGCNN\footnotemark[2]} ) & \multicolumn{2}{c}{$\qquad\ \ \ \qquad$83.03$\qquad\qquad\ \ $=$\qquad\qquad\qquad\ $\textbf{83.03}$\qquad\qquad\qquad$} \\
				\bottomrule[1pt]
		\end{tabular}	}
	\end{center}
	\caption{Comparisons of 3D shape classification accuracy between different methods on the ModelNet40 dataset. \textbf{NR/NR} denotes that neural networks were learned and tested with \textbf{N}o \textbf{R}otations. \textbf{NR/AR} denotes that neural networks were learned with \textbf{N}o \textbf{R}otations and tested with \textbf{A}rbitrary \textbf{R}otations. Experimental results show that the REQNN exhibited the highest rotation robustness. Note that the classification accuracy of the REQNN in scenarios of NR/NR and NR/AR was the same due to the rotation-equivariance property of the REQNN.}
	\label{tab:results_comparison}
\end{table}

\section{Experiments}\label{sec:exp}

Properties of the rotation equivariance and the permutation invariance of REQNNs could be proved theoretically, please see our supplementary materials for details. In order to demonstrate other advantages of REQNNs, we conducted the following experiments. We revised three widely used neural networks to REQNNs for the shape classification task, including PointNet++ \cite{qi2017pointnet++}, DGCNN \cite{wang2018dynamic}, and PointConv \cite{wu2019pointconv}. We revised the PointNet \cite{qi2017pointnet} to a REQNN for the point cloud reconstruction task. In all experiments, we set $c=1$ in Equation~(\ref{eq:relu}) and set $\epsilon=10^{-5}$ in Equation~(\ref{eq:norm}).

\textbf{3D shape classification:} We used the ModelNet40 \cite{wu20153d} dataset (in this study, we used corresponding point clouds provided by PointNet \cite{qi2017pointnet}) and the 3D MNIST \cite{3dMNIST} dataset for shape classification. The ModelNet40 dataset consisted of 40 categories; and the 3D MNIST dataset consisted of 10 categories. Each shape consisted of 1024 points. In this experiment, we conducted experiments on three types of baseline neural networks, including (1) the original neural network learned without rotations, (2) the original neural network learned with the ${\rm z}$-axis rotations (the ${\rm z}$-axis rotations were widely used in \cite{qi2017pointnet++,wang2018dynamic,wu2019pointconv} for data augmentation), and (3) the REQNN learned without rotations (the REQNN naturally had the rotation-equivariance property, so it did not require any rotation augmentation). The testing set was generated by arbitrarily rotating each sample ten times. We will release this testing set when this paper is accepted.

As Table~\ref{tab:results_class}\footnote[3]{The classification accuracy in the scenario of NR/AR in Table~\ref{tab:results_class} and Table~\ref{tab:results_comparison} was slightly different for PointNet++ \cite{qi2017pointnet++} (23.57\% \emph{vs.} 21.35\%) and DGCNN \cite{wang2018dynamic} (30.05\% \emph{vs.} 29.74\%). It was because architectures of PointNet++\footnotemark[1] and DGCNN\footnotemark[2] examined in Table~\ref{tab:results_class} and Table~\ref{tab:results_comparison} were slightly different. Nevertheless, this did not essentially change our conclusions.} shows, the REQNN always outperformed all baseline neural networks learned with or without rotations. We achieved the highest accuracy of 83.03\% using the REQNN revised from DGCNN\footnotemark[2]. Baseline neural networks that were learned without rotations exhibited very low accuracy (21.93\%--31.34\% on the ModelNet40 dataset and 44.15\%--51.16\% on the 3D MNIST dataset). In comparison, baseline neural networks that were learned with ${\rm z}$-axis rotations had little improvement in rotation robustness.

Besides, we compared the REQNN with several state-of-the-art methods for 3D point cloud processing in two scenarios, including neural networks learned with \textbf{N}o \textbf{R}otations and tested with \textbf{N}o \textbf{R}otations, and neural networks learned with \textbf{N}o \textbf{R}otations and tested with \textbf{A}rbitrary \textbf{R}otations, as Table~\ref{tab:results_comparison} shows. Note that the classification accuracy of the REQNN in the scenario of NR/NR was the same as that of NR/AR, because the REQNN was rigorously rotation equivariant. The best REQNN in this paper (\emph{i.e.} the REQNN revised from the DGCNN\footnotemark[2]) achieved the highest accuracy of 83.03\% in the scenario of NR/AR, which indicated the significantly high rotation robustness of the REQNN. Traditional methods, including PointNet \cite{qi2017pointnet}, PointNet++ \cite{qi2017pointnet++}, Point2Sequence \cite{liu2019point2sequence}, KD-Network \cite{klokov2017escape}, RS-CNN \cite{liu2019relation}, and DGCNN \cite{wang2018dynamic}, achieved high accuracy in the scenario of NR/NR. However, these methods performed poor in the scenario of NR/AR, because they could not deal with point clouds with unseen orientations. Compared with these methods, PRIN \cite{you2018prin} and QE-Capsule network \cite{zhao2019quaternion} made some progress in handling point clouds with unseen orientations. Our REQNN outperformed them by 14.18\% and 8.96\%, respectively, in the scenario of NR/AR.

\textbf{3D point cloud reconstruction:} In this experiment, we aimed to prove that we could rotate intermediate-layer quaternion features of the original point cloud to synthesize new point clouds with target orientations. Therefore, we learned a REQNN revised from the PointNet \cite{qi2017pointnet} for point cloud reconstruction on the ShapeNet \cite{chang2015shapenet} dataset. Each point cloud consisted of 1024 points in our implementation. We took the output quaternion feature of the top fourth linear transformation layer of the REQNN to synthesize quaternion features with different orientations. Such synthesized quaternion features were used to reconstruct point clouds with target orientations.

\begin{figure*}[tbp]
	\begin{center}
		\includegraphics[width=\linewidth]{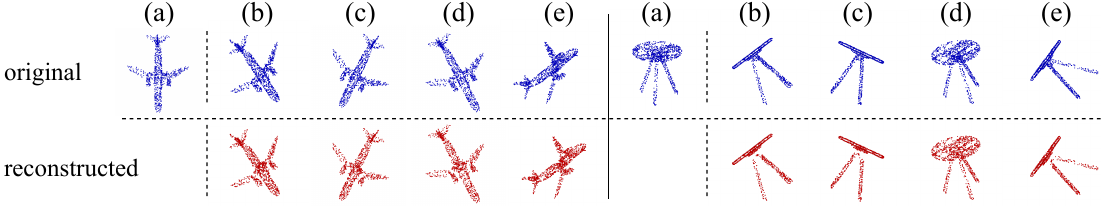}
	\end{center}
	\caption{Manual manipulation of intermediate-layer features to control the object rotation in 3D point cloud reconstruction. The experiment was conducted to prove that point clouds reconstructed using the synthesized quaternion features had the same orientations as point clouds generated by directly rotating the original point cloud. Here we displayed results of four random orientations for each point cloud. Point clouds (``original'' (b-e)) were generated by directly rotating the original point cloud (``original'' (a)) around axis {\small$[0.46,0.68,0.56]^{\top}$} with angle $\frac{\pi}{3}$, around axis {\small$[-0.44,-0.61,0.66]^{\top}$} with angle $\frac{\pi}{4}$, around axis {\small$[0.34,0.94,0.00]^{\top}$} with angle $\frac{\pi}{6}$, and around axis {\small$[0.16,0.83,0.53]^{\top}$} with angle $\frac{2\pi}{3}$, respectively. Given a specific intermediate-layer quaternion feature of the original point cloud (``original'' (a)), we rotated the quaternion feature with the same angles to obtain quaternion features with different orientations, which were used to reconstruct point clouds (``reconstructed'' (b-e)).}
	\label{fig:reconstruction}
\end{figure*}

As Fig.~\ref{fig:reconstruction} shows, for each given point cloud (Fig.~\ref{fig:reconstruction} ``original'' (a)), we directly rotated it with different angles (Fig.~\ref{fig:reconstruction} ``original'' (b-e)). For comparison, we rotated the corresponding quaternion feature of the original point cloud with the same angles to synthesize quaternion features. These generated quaternion features were used to reconstruct point clouds (Fig.~\ref{fig:reconstruction} ``reconstructed'' (b-e)). We observed that these reconstructed point clouds had the same orientations with those of point clouds generated by directly rotating the original point cloud.

\section{Conclusion}
In this paper, we have proposed a set of generic rules to revise various neural networks for 3D point cloud processing to REQNNs. We have theoretically proven that the proposed rules can ensure each layerwise operation in the neural network is rotation equivariant and permutation invariant. Experiments on various tasks have shown the rotation robustness of REQNNs.

We admit that revising a neural network to a REQNN has some negative effects on its representation capacity. Besides, it is challenging to revise all layerwise operations in all neural networks for 3D point cloud processing.

~\\
\noindent
\textbf{Acknowledgments}  The work is partially supported by the National Key Research and Development Project (No. 213), the National Nature Science Foundation of China (No. 61976160, U19B2043, and 61906120), the Special Project of the Ministry of Public Security (No. 20170004), and the Key Lab of Information Network Security, Ministry of Public Security (No.C18608).

\bibliographystyle{splncs04}
\bibliography{egbib}

\newpage
\appendix
\section{Quaternion Operations}

A quaternion {$\mathbf{q}=q_0+q_1\boldsymbol{i}+q_2\boldsymbol{j}+q_3\boldsymbol{k}\in\mathbb{H}$} is a hyper-complex number with a real part ({$q_0$}) and three imaginary parts ({$q_1\boldsymbol{i},q_2\boldsymbol{j},q_3\boldsymbol{k}$}), where {$q_0,q_1,q_2,q_3\in\mathbb{R}$}; $\mathbb{H}$ denotes the algebra of quaternions. The products of basis elements $\boldsymbol{i}$, $\boldsymbol{j}$, and $\boldsymbol{k}$ are defined by $\boldsymbol{i}^2=\boldsymbol{j}^2=\boldsymbol{k}^2=\boldsymbol{i}\boldsymbol{j}\boldsymbol{k}=-1$ and $\boldsymbol{i}\boldsymbol{j}=\boldsymbol{k}, \boldsymbol{j}\boldsymbol{k}=\boldsymbol{i},\boldsymbol{k}\boldsymbol{i}=\boldsymbol{j},\boldsymbol{j}\boldsymbol{i}=-\boldsymbol{k},\boldsymbol{k}\boldsymbol{j}=-\boldsymbol{i}$, and $\boldsymbol{i}\boldsymbol{k}=-\boldsymbol{j}$.

Just like complex numbers, given two quaternions $\mathbf{p}=p_0+p_1\boldsymbol{i}+p_2\boldsymbol{j}+p_3\boldsymbol{k}$ and $\mathbf{q}=q_0+q_1\boldsymbol{i}+q_2\boldsymbol{j}+q_3\boldsymbol{k}$,
a series of quaternion operations can be defined as follows.

\textbf{Addition:} $\mathbf{p}+\mathbf{q}=(p_0+q_0)+(p_1+q_1)\boldsymbol{i}+(p_2+q_2)\boldsymbol{j}+(p_3+q_3)\boldsymbol{k}$.

\textbf{Scalar multiplication:} $\lambda \mathbf{q}=\lambda q_0+\lambda q_1\boldsymbol{i}+\lambda q_2\boldsymbol{j}+\lambda q_3\boldsymbol{k}$.

\textbf{Element multiplication:} $\mathbf{p}\mathbf{q}=(p_0q_0-p_1q_1-p_2q_2-p_3q_3)+(p_0q_1+p_1q_0+p_2q_3-p_3q_2)\boldsymbol{i}+(p_0q_2-p_1q_3+p_2q_0+p_3q_1)\boldsymbol{j}+(p_0q_3+p_1q_2-p_2q_1+p_3q_0)\boldsymbol{k}$.

Note that multiplication of two quaternions is non-commutative, because $\mathbf{q}\mathbf{p}=(q_0p_0-q_1p_1-q_2p_2-q_3p_3)+(q_0p_1+q_1p_0+q_2p_3-q_3p_2)\boldsymbol{i}+(q_0p_2-q_1p_3+q_2p_0+q_3p_1)\boldsymbol{j}+(q_0p_3+q_1p_2-q_2p_1+q_3p_0)\boldsymbol{k}\neq \mathbf{p}\mathbf{q}$.

\textbf{Norm:} $\lVert\mathbf{q} \rVert$=$\sqrt{\mathbf{q}\overline{\mathbf{q}}}=\sqrt{q_0^2+q_1^2+q_2^2+q_3^2}$, where $\overline{\mathbf{q}}= q_0- q_1\boldsymbol{i}- q_2\boldsymbol{j}- q_3\boldsymbol{k}$ is the conjugation of $\mathbf{q}$.

\section{From Layerwise Rotation-Equivariance Property to the Rotation-Equivariance Property of the REQNN}
Let {$\mathbf{x}\in\mathbb{H}^{n}$} and $\mathbf{y}=\mathbf{\Phi}(\mathbf{x})=\Phi_L(\Phi_{L-1}(\cdots\Phi_1(\mathbf{x})))\in\mathbb{H}^C$ denote the input and the output of the REQNN, respectively. Let $\mathbf{f}_l=\Phi_l(\mathbf{f}_{l-1})\in\mathbb{H}^d$ denote the output of the {$l$}-th layer. We prove that the layerwise rotation equivariance can ensure the rotation-equivariance property of the REQNN. \emph{I.e.} if $\Phi_l(\mathbf{f}_{l-1}^{(\theta)})=\mathbf{R}\circ \Phi_l(\mathbf{f}_{l-1})\circ\overline{\mathbf{R}}$, \emph{s.t.} $\mathbf{f}_{l-1}^{(\theta)}\triangleq \mathbf{R}\circ \mathbf{f}_{l-1}\circ\overline{\mathbf{R}}$, $\forall l\in[1,2,...,L]$, then $\mathbf{\Phi}(\mathbf{x}^{\theta}) = \mathbf{R}\circ \mathbf{y} \circ\overline{\mathbf{R}} $.

\begin{equation}
\begin{aligned}
\mathbf{\Phi}(\mathbf{x}^{\theta}) =&\Phi_L(\Phi_{L-1}(\cdots  \Phi_1(\mathbf{R}\circ\mathbf{x}\circ\overline{\mathbf{R}}))) \\
=&\Phi_L(\Phi_{L-1}(\cdots \mathbf{R}\circ \Phi_1(\mathbf{x})\circ\overline{\mathbf{R}})) \\
& \dots \\
=&\Phi_L(\mathbf{R}\circ\Phi_{L-1}(\cdots  \Phi_1(\mathbf{x}))\circ\overline{\mathbf{R}}) \\
=&\mathbf{R}\circ\Phi_L(\Phi_{L-1}(\cdots  \Phi_1(\mathbf{x})))\circ\overline{\mathbf{R}} \\
=&\mathbf{R}\circ \mathbf{y} \circ\overline{\mathbf{R}}
\end{aligned}
\end{equation}

\section{Proofs of Layerwise Rotation Equivariance}

To prove that a layerwise operation $\Phi(\cdot)$ is rotation equivariant, $\Phi(\cdot)$ should satisfy $\Phi(\mathbf{f}^{(\theta)})=\mathbf{R}\circ \Phi(\mathbf{f})\circ\overline{\mathbf{R}}$, where $\mathbf{f}^{(\theta)}$ indicates the quaternion feature that was obtained by rotating $\mathbf{f}$ around an axis $\mathbf{o}=o_1\boldsymbol{i}+o_2\boldsymbol{j}+o_3\boldsymbol{k}$ with an angle $\theta$. Such a rotation can be represented using quaternion $\mathbf{R}=\cos\frac{\theta}{2}+\sin\frac{\theta}{2}(o_1\boldsymbol{i}+o_2\boldsymbol{j}+o_3\boldsymbol{k})$ and its conjugation $\overline{\mathbf{R}}=\cos\frac{\theta}{2}-\sin\frac{\theta}{2}(o_1\boldsymbol{i}+o_2\boldsymbol{j}+o_3\boldsymbol{k})$, \emph{i.e.} $\mathbf{f}^{(\theta)}= \mathbf{R}\circ \mathbf{f}\circ\overline{\mathbf{R}}$. We only use symbol $\theta$ to represent a rotated quaternion feature and omit the symbol $\mathbf{o}$ for convenience.

\subsection{Convolution Operation}\label{appendix:proof_conv_rotation}

Let $\mathbf{f}=[\mathbf{f}_{1},\dots,\mathbf{f}_{d} ]^{\top}\in\mathbb{H}^d$ denote the quaternion feature of a point in the point cloud. Note that 3D point cloud processing usually uses the specific convolution with 1$\times$1 kernels. Let us take this specific convolution as the example to prove that the revised convolution operation is rotation equivariant. The revised convolution operation,
\begin{equation}
\begin{aligned}
Conv(\mathbf{f}) &= w\otimes\mathbf{f} \\
&=\begin{bmatrix}
w_{11} &   \cdots & w_{1d}      \\
\vdots & \ddots & \vdots \\
w_{D1}    & \cdots & w_{Dd}
\end{bmatrix}\otimes
\begin{bmatrix}
\mathbf{f}_{1}  \\
\vdots  \\
\mathbf{f}_{d}
\end{bmatrix} \\
& = \begin{bmatrix}
(w_{11}  \mathbf{f}_{1}  +  \dots + w_{1d}   \mathbf{f}_{d} )  \\
\vdots  \\
(w_{D1}  \mathbf{f}_{1}  +  \dots + w_{Dd}   \mathbf{f}_{d} )
\end{bmatrix},
\end{aligned}
\end{equation}
is rotation-equivariant, because $Conv(\mathbf{f}^{(\theta)})=\mathbf{R}\circ Conv(\mathbf{f}) \circ \overline{\mathbf{R}}$. The proof is given as follows.
\begin{equation}
\begin{aligned}
Conv(\mathbf{f}^{(\theta)})&=w\otimes(\mathbf{R}\circ \mathbf{f} \circ \overline{\mathbf{R}} )\\
&=\begin{bmatrix}
w_{11} &   \cdots & w_{1d}      \\
\vdots & \ddots & \vdots \\
w_{D1}    & \cdots & w_{Dd}
\end{bmatrix}\otimes (\mathbf{R}\circ
\begin{bmatrix}
\mathbf{f}_{1}  \\
\vdots \\
\mathbf{f}_{d}
\end{bmatrix} \circ \overline{\mathbf{R}})\\
& = \begin{bmatrix}
(w_{11}  (\mathbf{R}\mathbf{f}_{1}\overline{\mathbf{R}})  +  \dots + w_{1d}   (\mathbf{R}\mathbf{f}_{d} \overline{\mathbf{R}})) \\
\vdots \\
(w_{D1}  (\mathbf{R}\mathbf{f}_{1}\overline{\mathbf{R}})  +  \dots + w_{Dd}   (\mathbf{R}\mathbf{f}_{d} \overline{\mathbf{R}}))
\end{bmatrix} \\
& = \begin{bmatrix}
(\mathbf{R} (w_{11} \mathbf{f}_{1})\overline{\mathbf{R}}  +  \dots + \mathbf{R} (w_{1d}   \mathbf{f}_{d} )\overline{\mathbf{R}}) \\
\vdots  \\
(\mathbf{R}(w_{D1} \mathbf{f}_{1})\overline{\mathbf{R}} +  \dots +  \mathbf{R}(w_{Dd}  \mathbf{f}_{d} )\overline{\mathbf{R}})
\end{bmatrix} \\
& = \mathbf{R}\circ (w\otimes\mathbf{f} )\circ \overline{\mathbf{R}} \\
& = \mathbf{R}\circ Conv(\mathbf{f} )\circ \overline{\mathbf{R}}.
\end{aligned}
\end{equation}

\subsection{ReLU Operation}

The revised ReLU operation, $ReLU(\mathbf{f}_v) = \frac{\lVert\mathbf{f}_v\rVert}{\max\{\lVert\mathbf{f}_v\rVert,c\}} \mathbf{f}_v$, is rotation-equivariant, because $ReLU(\mathbf{f}_v^{(\theta)})=\mathbf{R} ReLU(\mathbf{f}_v)  \overline{\mathbf{R}}$. The proof is given as follows.
\begin{equation}
\begin{aligned}
ReLU(\mathbf{f}_v^{(\theta)}) &=\frac{\lVert \mathbf{R} \mathbf{f}_v  \overline{\mathbf{R}} \rVert}{\max\{\lVert\mathbf{R}\mathbf{f}_v  \overline{\mathbf{R}}\rVert,c\}} (\mathbf{R} \mathbf{f}_v  \overline{\mathbf{R}})\\
&= \frac{\lVert  \mathbf{f}_v  \rVert}{\max\{\lVert\mathbf{f}_v \rVert,c\}} (\mathbf{R} \mathbf{f}_v  \overline{\mathbf{R}})\\
&= \mathbf{R} (\frac{\lVert  \mathbf{f}_v  \rVert}{\max\{\lVert\mathbf{f}_v \rVert,c\}}  \mathbf{f}_v)  \overline{\mathbf{R}}\\
&= \mathbf{R} ReLU(\mathbf{f}_v) \overline{\mathbf{R}}.
\end{aligned}
\end{equation}
Note that $\lVert \mathbf{R} \mathbf{f}_v  \overline{\mathbf{R}} \rVert = \lVert \mathbf{f}_v \rVert$, because rotating a quaternion will not change its norm.

\subsection{Batch-Normalization}
\begin{sloppypar}
The revised batch-normalization, $norm(\mathbf{f}_v^{(i)})=\frac{\mathbf{f}_v^{(i)}}{ \sqrt{\mathbb{E}_j[\lVert \mathbf{f}_v^{(j)}\rVert^2]+ \epsilon}}$, is rotation-equivariant, because $norm((\mathbf{f}_v^{(i)})^{(\theta)})=\mathbf{R} norm(\mathbf{f}_v^{(i)}) \overline{\mathbf{R}}$. The proof is given as follows.
\end{sloppypar}
\begin{equation}
\begin{aligned}
norm((\mathbf{f}_v^{(i)})^{(\theta)}) &= \frac{\mathbf{R}\mathbf{f}_v^{(i)}\overline{\mathbf{R}}}{ \sqrt{\mathbb{E}_j[\lVert \mathbf{R}\mathbf{f}_v^{(j)}\overline{\mathbf{R}}\rVert^2]+ \epsilon}} \\
&= \frac{\mathbf{R}\mathbf{f}_v^{(i)}\overline{\mathbf{R}}}{ \sqrt{\mathbb{E}_j[\lVert \mathbf{f}_v^{(j)}\rVert^2]+ \epsilon}} \\
&= \mathbf{R}\frac{\mathbf{f}_v^{(i)}}{ \sqrt{\mathbb{E}_j[\lVert \mathbf{f}_v^{(j)}\rVert^2]+ \epsilon}} \overline{\mathbf{R}}\\
&= \mathbf{R} norm(\mathbf{f}_v^{(i)})  \overline{\mathbf{R}}.
\end{aligned}
\end{equation}

\subsection{Max-Pooling Operation}

\begin{sloppypar}
The revised max-pooling operation, $\textit{maxPool}({\bf f})= {\bf f}_{\hat{v}}$, ${\rm s.t.}\ \hat{v}=\mathop{\arg\max}_{v=1,\dots,d.}[\Vert{\bf f}_v\Vert]$,
is rotation-equivariant, because $maxPool(\mathbf{f}^{(\theta)})=\mathbf{R} maxPool(\mathbf{f})  \overline{\mathbf{R}}$. The proof is given as follows.
\end{sloppypar}
\begin{equation}
\begin{aligned}
maxPool(\mathbf{f}^{(\theta)}) &= \mathbf{R}{\bf f}_{\hat{v}}\overline{\mathbf{R}}\quad{\rm s.t.}\quad \hat{v}=\mathop{\arg\max}_{v=1,\dots,d.}[\Vert\mathbf{R}{\bf f}_v\overline{\mathbf{R}}\Vert]\\
&=\mathbf{R}{\bf f}_{\hat{v}}\overline{\mathbf{R}}\quad \hat{v}=\mathop{\arg\max}_{v=1,\dots,d.}[\Vert{\bf f}_v\Vert]\\
&= \mathbf{R} \textit{maxPool}({\bf f}) \overline{\mathbf{R}}.
\end{aligned}
\end{equation}

\section{Element-Wise Max-Pooling Operation}
In point cloud processing, a special element-wise max-pooling operation is widely used for aggregating a set of neighboring points' features into a local feature. Let $ f \in\mathbb{R}^{D\times K}$ denote the features of $K$ neighboring points. Each element of $f$, \emph{i.e.} $f_k\in\mathbb{R}^D$, denotes the feature of a specific neighboring point. Let $f^{\rm upper}\in\mathbb{R}^D$ denote the output local feature. The element-wise max-pooling operation is formulated as follows.
	\begin{equation}\label{eq:max}
	\begin{aligned}
	\mathop{\mathbf{MAX}}(f)&=\mathop{\mathbf{MAX}}
	\begin{bmatrix}
	f_{11}      & \cdots & f_{1K}      \\
	\vdots & \ddots & \vdots \\
	f_{D1}      & \cdots & f_{DK}
	\end{bmatrix} \\
	& \xlongequal{\textbf{define}}
	<\max\limits_{k=1,\dots,K}f_{1k},\dots,\max\limits_{k=1,\dots,K}f_{Dk}>^{\top}\\
	\end{aligned}
	\end{equation}
\begin{sloppypar}
To extend this special max-pooling operation to be suitable for quaternion operations, we replace each max operation $\max\limits_{k=1,\dots,K}f_{dk}$ in Equation~(\ref{eq:max}) by $\max\limits_{k=1,\dots,K}[\Vert{\bf f}_{dk}\Vert]$, where $\mathbf{f}_{dk}\in\mathbb{H}$ is a quaternion.
\end{sloppypar}

\end{document}